\documentclass[10pt]{amsart}

\usepackage[american]{babel}

\usepackage{amsmath}
\usepackage{amssymb}
\usepackage{bm}
\usepackage{mathtools}
\usepackage{stmaryrd}
\usepackage{calrsfs}

\usepackage{graphicx}
\usepackage{psfrag}
\usepackage{float}

\usepackage{color}
\usepackage{xcolor}

\usepackage{listings}

\definecolor{keywordcolor}{rgb}{0.2, 0.2, 0.75}
\definecolor{stringcolor}{rgb}{0.0, 0.5, 0.0}
\definecolor{commentcolor}{rgb}{0.5, 0.5, 0.5}
\definecolor{backgroundcolor}{rgb}{0.95, 0.95, 0.95}
\definecolor{numbercolor}{rgb}{0.3, 0.3, 0.3}

\lstdefinestyle{pythonstyle}{
    language=Python,
    backgroundcolor=\color{backgroundcolor},
    basicstyle=\ttfamily\small,
    keywordstyle=\color{keywordcolor}\bfseries,
    stringstyle=\color{stringcolor},
    commentstyle=\color{commentcolor}\itshape,
    numberstyle=\tiny\color{numbercolor},
    numbers=left,
    numbersep=10pt,
    showstringspaces=false,
    breaklines=true,
    frame=single,
    tabsize=4,
    captionpos=b
}

\definecolor{felix}{rgb}{1,0,0}

\definecolor{marly}{rgb}{1,0,0.4}

\definecolor{joseph}{rgb}{0,0,1}

\definecolor{anna}{HTML}{d595fc}

\definecolor{arav}{rgb}{0.037,0.43,0.631}

\definecolor{bryan}{HTML}{fc9221}

\definecolor{jason}{HTML}{4a8e1f}

\definecolor{jiya}{HTML}{4c2dbd}

\usepackage{tikz}
\usetikzlibrary{matrix,positioning}
\usepackage{tikz-cd}

\usepackage{pgfplots}
\pgfplotsset{compat=1.18}
\usepgfplotslibrary{colormaps}

\usepackage{xy}
\input xy
\xyoption{all}

\usepackage{algpseudocode}

\usepackage{newunicodechar}
\newunicodechar{≈}{\approx}

\usepackage[
  backend=biber,
  style=numeric,
  sorting=nyt,
  giveninits=true,
  maxnames=3, minnames=1,
  doi=true, url=true, isbn=true, eprint=true,
  urldate=short
]{biblatex}

\DeclareNameAlias{default}{given-family}
\DeclareFieldFormat{labeldate}{\mkbibparens{#1}}
\DeclareFieldFormat{date}{#1}

\newbibmacro*{date+extrayear}{
  \iffieldundef{labelyear}
    {}
    {\printtext[date]{\printlabelyear}}}

\renewbibmacro*{author+date}{%
  \printnames{author}%
  \setunit{\addspace}%
  \printtext[date]{\printdate}}

\usepackage{hyperref}
\hypersetup{colorlinks=true, linkcolor=blue, citecolor=magenta, urlcolor=green}
\usepackage{url}

\usepackage{fancyhdr}
\theoremstyle{definition}

\newtheorem*{remark*}{Remark}

\theoremstyle{remark}

\numberwithin{equation}{section}
\numberwithin{figure}{section}
\numberwithin{table}{section}

\keywords{Topic modeling, BERTopic, Transformer embeddings, Topic coherence}

\begin{document}

\mbox{}
\title{A Comparative Study of Transformer-Based\\
Embeddings for Topic Coherence}

\author{Alex Ding}
\address{Worcester Academy \\ Stanford Online High School}
\email{DSTworcester@163.com}

\author{Tarun Rapaka}
\address{Stanford Online High School}
\email{tarun.rapaka@gmail.com}

\author{Willy Rodriguez}
\address{Torus Actions}
\email{willyrv@gmail.com}

\author{Jason Yang}
\address{Lexington High School}
\email{jyang.super@gmail.com}

\date{\today}

\begin{abstract}
	Topic modeling is a branch of Natural Language Processing (NLP) that aims to organize large collections of texts into coherent groups according to word co-occurrence patterns, with Latent Dirichlet Allocation (LDA) remaining one of the most widely used and interpretable probabilistic approaches. Recent advances in NLP, particularly transformer-based language models, offer improved document representations. It is also known that the size of the model (in terms of number of parameters) has a significant impact in the performance of the language models on different pre-defined tasks. In this study, we systematically examine the effect of model size on topic quality by analyzing the performances of seven transformer-based language models (from small models such as MiniLM to large ones such as LLaMA-2) in a BERTopic pipeline on a variety of corpora. Topic quality is evaluated using coherence and divergence metrics following Röder et al. (2015). Our results indicate that model size, ranging from 22 million to 13 billion parameters, has a negligible impact on the quality of the topic, suggesting that smaller models can achieve comparable performance to larger models.
\end{abstract}

\bigskip
\maketitle

\bigskip

\section{Introduction}

Topic modeling is a central tool in Natural Language Processing (NLP) for revealing hidden semantic regularities in large corpora. In this view, a topic is modeled as a probability distribution over the vocabulary, while a document is represented as a mixture of several topics. The goal is to discover these recurring structures automatically so they can support downstream tasks such as summarization, document organization, and corpus analysis. Classical approaches, such as Latent Dirichlet Allocation (LDA) \cite{blei2003latent} and Non‑negative Matrix Factorization (NMF) \cite{lee1999learning}, generally rely on a bag‑of‑words representation that ignores word order and treats documents as token‑frequency vectors. While fast and often interpretable, such models struggle with context and polysemy, which can yield topics of limited coherence, especially on specialized or heterogeneous data.

Recent progress mitigates these issues by using transformer‑based architectures, notably BERT (Bidirectional Encoder Representations from Transformers) \cite{devlin2019bert}. Unlike earlier pipelines, BERT produces context‑dependent embeddings, so a word’s representation adapts to its surrounding tokens. These richer encodings have advanced many applications, from text classification to translation and question answering.

Building on this foundation, BERTopic \cite{grootendorst2022bertopic} is a recently developed and highly flexible framework for topic modeling that leverages transformer-based language models. It operates in three main stages: (1) creating document embeddings using a pre-trained language model, (2) clustering those embeddings into groups of semantically similar documents, and (3) extracting topic representations using a class-based variation of TF-IDF, which identifies representative keywords for each cluster. This modular and embedding-based approach overcomes many of the limitations of traditional models and yields more coherent, interpretable, and domain-adaptive topics.

While BERTopic relies on document-level embeddings combined with clustering, subsequent work has explored richer modeling strategies that incorporate contextual information directly at the word level. In particular, the Contextualized Word Topic Model (CWTM) \cite{fang2024cwtm} replaces the bag-of-words assumption with BERT-based embeddings for individual tokens, allowing each word to be represented according to its surrounding context. This formulation enables the model to distinguish semantically different usages of the same word and to associate topic distributions with specific word occurrences. Empirical evaluations show that such approaches improve topic coherence and diversity, while also demonstrating robustness to out-of-vocabulary words compared to traditional probabilistic models.

More recently, neural topic modeling approaches have explored tighter integration between embedding spaces and word distributions. For instance, FASTopic \cite{wu2024fastopic} introduces a fast and stable framework that leverages pretrained Transformer-based document, topic, and word embeddings to reconstruct document-word distributions. Its core mechanism relies on entropy-regularized optimal transport to model document-word relationships, which helps mitigate issues such as topic redundancy and collapse. Empirical results show that such approaches can achieve competitive or improved topic quality (in terms of coherence and diversity), while significantly reducing training time and improving robustness to hyperparameter choices.

Nevertheless, interpreting the extracted topics remains a challenging task. By interpretability, we refer to the ability of a human reader to make sense of a topic based on its most representative words. For example, consider a topic represented by the keywords: \textit{infection, symptoms, treatment, fever, diagnosis}. A human observer would likely interpret this as a health-related topic concerning infectious diseases. However, if the representative words are vague or too general (e.g., \textit{thing, process, case, information}), the topic becomes difficult to interpret. This illustrates the importance of producing interpretable topics in downstream tasks. This is particularly important for applications that require human-in-the-loop validation or exploration of thematic structures. While some metrics have been proposed to quantify the interpretability and overall quality of the topics, this remains an open research question. A metric is generally considered reliable if it correlates well with human judgments regarding how coherent or meaningful a topic appears \cite{roder2015exploring}.

Contextual embeddings capture semantic similarity. However, they are not always well-suited for forming distinct and coherent clusters, which are essential for high-quality topic extraction. To address this issue, DeTiME \cite{xu-etal-2023-detime} introduces a hybrid framework that combines encoder–decoder large language models, variational autoencoders, and diffusion models. The approach first fine-tunes a transformer model on a paraphrasing task to produce embeddings that are more amenable to clustering. These representations are then compressed and modeled through a VAE that jointly reconstructs both bag-of-words signals and embedding features. Finally, a diffusion process is applied in the embedding space to refine latent representations and improve the coherence of generated topics. Experimental results show that this approach significantly improves clustering performance and topic quality, while also enabling more fluent and interpretable topic-guided text generation.

Transformer-based language models have revolutionized the NLP field, offering powerful and context-aware word representations. These representations are particularly effective for tasks such as topic decomposition, where understanding semantic similarity is key. BERTopic leverages this strength by allowing users to plug in different transformer-based models at the first step of its pipeline, offering high flexibility in adapting to various resource constraints and domains.

A common observation in NLP is that increasing the size of a language model (i.e. the number of parameters) typically leads to improved accuracy and task performance \cite{kaplan2020scaling}. While this trend is well established for tasks such as classification or translation, there has been limited exploration into how model size affects the interpretability and coherence of topics produced in topic modeling tasks. This raises a critical question: How does the choice of language model, particularly its size (measured in number of parameters), affect the quality of the extracted topics? Larger language models often yield better results across NLP benchmarks, but they come with higher computational costs. On the other hand, smaller models offer practical advantages in terms of speed and efficiency, making them attractive in resource-constrained settings. Understanding this trade-off is essential for guiding the selection of language models in topic modeling workflows.

In this study, we used a pipeline based on BERTopic to investigate how the quality of topic decomposition is influenced by the size of the underlying language model. We begin by benchmarking seven pre-trained transformer-based models on the widely used 20 Newsgroups dataset, which contains roughly 20,000 documents categorized into 20 distinct newsgroups. This provides a standard baseline for evaluating topic coherence and alignment with human-labeled categories.

We use the same method to process ten more datasets, including an in-house customized database selected from PubMed composed by all publications having the words "Large Language Models", "AI-assisted diagnosis" or "Artificial Intelligence" in their abstracts or content between January 2024 and May 2025. This dataset containing the most current trend in biomedical research on AI and LLM is open and can be freely downloaded \cite{pubmed-dataset}. The experimental code for this paper has also been open sourced and can be found at \cite{ml2nlp-code}.

To objectively evaluate topic models, we take advantage of standardized measures. To evaluate the interpretability, coherence and semantic consistency of topics and to discriminate topic models, we measure topic coherence and topic divergence in BERTopic. We also systemically compare various models with each other and therefore pick the best model to fit our text document.

The remainder of the paper is organized as follows: Section~\ref{sec:background} reviews required background and metrics; Section~\ref{sec:methods} details datasets, preprocessing, and the evaluation pipeline; Section~\ref{sec:results} presents the experiments and findings and Section~\ref{sec:conclusions} discusses implications and future directions.

\section{Background}~\label{sec:background}
In this section we gather the core concepts needed for the experiments. We begin by describing two well known topic‑modeling methods: Latent Dirichlet Allocation (LDA) and Non‑negative Matrix Factorization (NMF). These methods have been widely used for extracting hidden themes from document collections. We then summarize how these models operate and where their limitations lie, and we introduce two evaluation metrics for topics: \emph{coherence} and \emph{divergence}, used to compare quality and separation among topics. Finally, we give a concise overview of the Transformer architecture that underpins modern language models and serves as the backbone of BERTopic used in this work.

\subsection{Overview of NMF and LDA for topic modeling}






\paragraph{Non-negative Matrix Factorization (NMF).}
Proposed by Lee and Seung \cite{lee1999learning}, NMF is a linear‑algebraic approach widely adopted in topic modeling for uncovering structured patterns in non‑negative data. In NLP it is applied to a non‑negative document–term matrix \( V \in \mathbb{R}_{\ge 0}^{m\times n} \), where \( m \) denotes the number of documents and \( n \) the vocabulary size. The objective is to approximate \( V \) by a product of two lower‑rank non‑negative matrices:
\[
V \approx WH
\]
where \( W \in \mathbb{R}_{\ge 0}^{m\times k} \) links documents to topics and \( H \in \mathbb{R}_{\ge 0}^{k\times n} \) stores topic–word weights. Non‑negativity encourages sparse, parts‑based decompositions that are easy to read: topics emphasize distinct sets of high‑weight terms, and documents become mixtures of those topics. In contrast to generative models such as LDA, NMF is deterministic and algebraic, typically fast and interpretable—especially when the input uses TF–IDF.

\paragraph{Latent Dirichlet Allocation (LDA).}
LDA \cite{blei2003latent} is a foundational probabilistic model that aims to infer latent topics from observed word usage, ideally yielding semantically meaningful themes (e.g., “politics” or “sports”). It posits that each document is a mixture of topics and that each topic defines a probability distribution over words drawn from a fixed vocabulary. 

Formally, for each document a multinomial topic proportion is drawn from a Dirichlet prior; for every word position, a topic is sampled from that document’s proportions, and the word token is then sampled from the selected topic’s word distribution. This hierarchical Bayesian construction encodes that documents mix several topics in different proportions and that words associated with a topic tend to co‑occur.

The steps that LDA goes through to generate topics are as follows:
\begin{enumerate}
    \item For each topic \( k = 1, 2, \dots, K \), draw a distribution over words: \( \phi_k \sim \text{Dir}(\beta) \)
    \item For each document \( d \) in the corpus:
    \begin{enumerate}
        \item Draw topic proportions: \( \theta_d \sim \text{Dir}(\alpha) \)
        \item For each word \( w_{dn} \):
        \begin{enumerate}
            \item Draw a topic assignment: \( z_{dn} \sim \text{Multinomial}(\theta_d) \)
            \item Draw the word: \( w_{dn} \sim \text{Multinomial}(\phi_{z_{dn}}) \)
        \end{enumerate}    
    \end{enumerate}
\end{enumerate}

In this formulation, \(\alpha\) and \(\beta\) are the Dirichlet hyperparameters controlling, respectively, the sparsity of document–topic proportions and topic–word distributions \cite{niu2015visual}. LDA inference aims to compute the posteriors of the latent variables \((\theta,\phi,z)\) conditioned on the observed words. Because an exact solution is infeasible, one resorts to approximations such as variational inference \cite{blei2003latent} or collapsed Gibbs sampling \cite{griffiths2004finding}.

\begin{figure}[ht]
    \centering
    \includegraphics[width=0.95\textwidth]{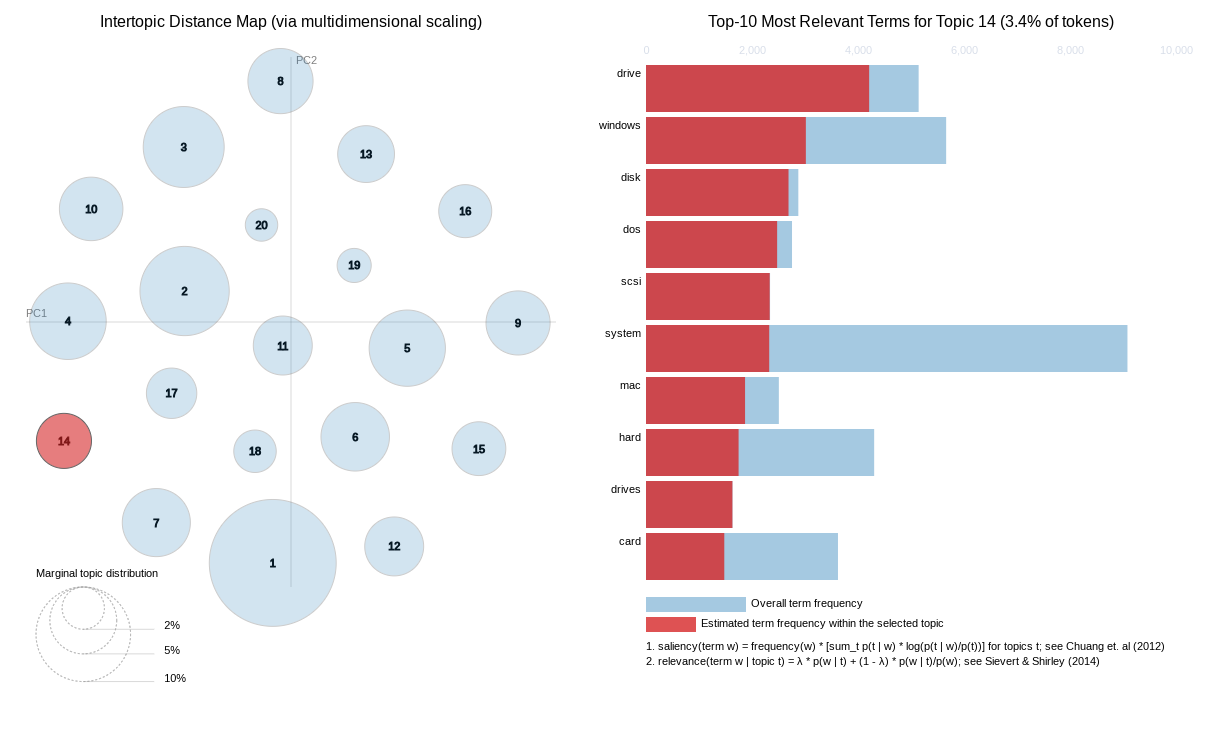}
    \caption{Visualization of the LDA topic decomposition with 20 topics trained on the \textit{20 Newsgroups} dataset.}
    \label{fig:lda_visualization}
\end{figure}

The model outputs per‑topic word distributions and, for each document, a mixture over topics—useful for summarization, clustering, and exploratory analysis. As an illustration, Fig.~\ref{fig:lda_visualization} visualizes a 20‑topic LDA trained on the \textit{20 Newsgroups} corpus. \textbf{Left:} an inter‑topic distance map produced by multidimensional scaling (MDS); each circle denotes a topic and its size reflects its prevalence in the corpus, with Topic~14 highlighted. \textbf{Right:} the ten most relevant terms for Topic~14, computed via a relevance score that balances frequency and distinctiveness. Red bars indicate estimated within‑topic term frequencies, whereas blue bars show corpus‑level frequencies. The visualization follows \cite{sievert2014ldavis} and uses relevance as in \cite{chuang2012termite}.

\subsection{Word embeddings and contextual representations}
Earlier NLP pipelines typically represented words as one‑hot indices, which ignore semantic relatedness and structural similarity. With the advent of word embeddings it became clear that mapping tokens into dense, low‑dimensional vectors captures both semantic proximity and syntactic relations more effectively.

A widely used method for learning such vectors is Word2Vec \cite{mikolov2013efficient}, which leverages local co‑occurrence statistics. Two training variants are common: \emph{CBOW}, which predicts a center word from its context, and \emph{Skip‑gram}, which predicts surrounding words from a target token. Despite their simplicity, these models yield high‑quality embeddings that encode syntactic and semantic regularities. For instance, if $emb(w)$ denotes the embedding of word $w$, one observes arithmetic patterns such as:

$$emb(\text{"king"}) - emb(\text{"man"}) + emb(\text{"woman"}) \approx emb(\text{"queen"})$$ which suggests that the space of the embedding catches some semantical relations that are present in natural language. 

Though both Word2Vec and similarly static embeddings (e. g., GloVe \cite{pennington2014glove}) allot one vector per word irrespective of context, they fail to disambiguate between a polysemous word (e. g.,“bank”as an institution versus a riverbank), hence the rise in popularity of contextual embeddings formed using deep neural models such as BERT.

Introduction of the \textbf{Transformer architecture} \cite{vaswani2017attention} changed the paradigm of language modeling from computing context free embeddings to using context sensitive ones. Unlike previous sequence models like RNN or LSTM that were able to leverage self attention mechanism and in turn the model was able to take into consideration the saliency of each word in a sentence when encoding another word, leading to better performance especially with respect to capturing long term dependencies, better word sense disambiguation, etc.

Let us take \textit{``He went to the bank to withdraw cash''} and \textit{``She sat by the bank of the river''} for example. In both sentences, the embedding representation of the word“bank”would not be the same because the transformer can utilize the attention mechanism to capture contextual clues, which then gives birth to the context-aware representations as the fundamental building blocks of modern pre-trained language models such as  BERT \cite{devlin2019bert}, RoBERTa \cite{liu2019roberta}, and GPT \cite{radford2018improving}, which have achieved groundbreaking results in many NLP applications.

\subsection{The Transformer Architecture}

Transformers \cite{vaswani2017attention} underpin most modern NLP systems. Unlike recurrent architectures (RNNs/LSTMs), they are built around \textbf{self‑attention}, enabling dependencies to be captured between any pair of tokens regardless of distance. A standard transformer comprises a stack of \textbf{encoders} and, for sequence‑to‑sequence tasks, a stack of \textbf{decoders}; each block contains a self‑attention layer and a position‑wise feed‑forward network.

\paragraph{Encoders}

Each encoder block combines multi‑head self‑attention with a feed‑forward network. Residual connections and layer normalization follow each sub‑layer to stabilize training.

\paragraph{Self-attention}

Given an input matrix $X \in \mathbb{R}^{N \times d}$ (with $N$ tokens and embedding dimension $d$), self‑attention lets every token attend to all others in the sequence.

\vspace{0.5em}
To capture diverse relations, self‑attention employs $n$ \textbf{heads}, each with its own learned projections:
\begin{itemize}
    \item $W_i^Q \in \mathbb{R}^{d \times d_K}$ (query projection),
    \item $W_i^K \in \mathbb{R}^{d \times d_K}$ (key projection),
    \item $W_i^V \in \mathbb{R}^{d \times d_K}$ (value projection).
\end{itemize}

Head $i$ computes its query, key, and value matrices as:
\begin{align*}
Q_i &= X \cdot W_i^Q, \\
K_i &= X \cdot W_i^K, \\
V_i &= X \cdot W_i^V.
\end{align*}

The head’s attention is computed via the scaled dot‑product:
\[
\text{Attention}_i = \text{softmax} \left(\frac{Q_i K_i^T}{\sqrt{d_K}} \right) \cdot V_i.
\]

This yields attention‑weighted representations indicating how strongly each token attends to others.

Outputs from the $n$ heads are concatenated and linearly projected:
\[
\text{MultiHead}(X) = \text{Concat}(A_1,A_2, \dots, A_n) \cdot W_O,
\]
where $W_O \in \mathbb{R}^{n d_K \times d}$.

\begin{remark*}
Self‑attention integrates evidence from the entire sequence. For instance, the embedding of “bank” shifts depending on whether nearby words imply finance or a riverside.
\end{remark*}

\paragraph{Feed-forward Neural Network}

After the attention layer, each token is passed independently through the same two-layer feed-forward network:
\[
\text{FFN}(x) = \text{ReLU}(xW_1 + b_1)W_2 + b_2.
\]

The nonlinearity equips the model to form richer token representations. The FFN operates position‑wise, i.e., the same network is applied independently to every token in the sequence.

\paragraph{Decoders}

Decoder blocks mirror encoders but add \textbf{encoder–decoder attention}, which lets the decoder query the encoder outputs so predictions can be conditioned on the source sequence. During training, masked self‑attention is used in the decoder to prevent peeking at future positions.

\paragraph{Final Softmax Application}

After the decoder stack, each output position yields a vector $v$. A final linear layer followed by a softmax converts $v$ into a probability distribution over the vocabulary:
\[
\hat{y} = \text{softmax}(W \cdot v + b).
\]

During inference, the model can generate words by selecting the most likely token (greedy decoding) or by sampling strategies. During training, loss functions such as cross-entropy or KL-divergence are used to compare predicted distributions against the ground truth, enabling optimization via backpropagation.

\subsection{Topic Coherence and Topic Divergence Metrics}

Evaluating the quality of topics produced by topic models is not an easy job, because most topic models are unsupervised, so we cannot use the common assessment metrics like accuracy or precision. Therefore, researchers propose special indexes, which evaluate the quality from two aspects—\textbf{topic coherence} and \textbf{topic divergence}.
The measurements can also tell us something about the interpretability and differentiation and meaning of the topics produced – almost always consistent with human judgements about how good topics are.

\textbf{Why are topic-coherence metrics useful?}
They provide a quantitative proxy for the \emph{interpretability} of topics, allowing practitioners to
(i) choose the number of topics or tune hyper-parameters without repeated human annotation,
(ii) compare different modelling algorithms on a common scale, and
(iii) track how thematic structure drifts over time.
Because these scores correlate strongly with human judgements, they have become a de-facto objective function when optimizing topic models~\cite{roder2015exploring}.

\subsubsection{Theoretical Framework for Topic Coherence}\label{paragraph:theoretical_coherence}

Following Röder et\,al.~\cite{roder2015exploring}, topic coherence is cast as a modular pipeline with four ingredients: \textbf{segmentation}, \textbf{probability estimation}, \textbf{confirmation measure}, and \textbf{aggregation}. This viewpoint yields a common mathematical template encompassing most coherence measures in the literature \cite{stevens2012}.

Let \( T = \{w_1, w_2, \dots, w_N\} \) be the top-\(N\) word list that represents a topic. \\

\textbf{Definition 2.1 Topic Coherence}:
Let \( S(T) = \{ (W'_i, W^*_i) \}_{i=1}^m \) be a segmentation of \( T \) into $m$ word-set pairs. Let \( P \colon \mathcal{W} \times \mathcal{W} \to [0,1] \) be a probability estimation function, and let \( M(W'_i, W^*_i; P) \in \mathbb{R} \) be a confirmation measure. Then the coherence of topic \( T \) is given by:
\[
Coherence(T) = \sum_{i=1}^m M(W'_i, W^*_i; P),
\]
where \( \Sigma \) is an aggregation operator (e.g., arithmetic mean).

\paragraph{Segmentation \( S \).}
Segmentation defines how the top-\( N \) words are split into meaningful subgroups. A common choice is the \textit{one-set segmentation}:
\[
S_{\text{one\_set}}(T) = \{ (\{w_i\}, T \setminus \{w_i\}) \mid w_i \in T \}.
\]
Other segmentation strategies include \emph{pairwise} segmentation (all unordered word pairs), or \emph{sliding window} based subsets.

\paragraph{Probability Estimation \( P \).}
The map \( P(w_i, w_j) \) quantifies how often \( w_i \) and \( w_j \) co‑occur. Common choices include:
\begin{itemize}
  \item \textbf{Document-based:} Proportion of documents in which both words co-occur
  \item \textbf{Sliding Window:} Frequency of co-occurence within a certain sized window (e. g., 70 tokens)
  \item \textbf{External Corpus:} Co-occurrence statistics based on very large reference corpora like Wikipedia
\end{itemize}

\paragraph{Confirmation Measure \( M \).}
The confirmation measure checks if the word set \( W' \) supports \( W^* \) at the level of \( P \) and can be represented as the following examples.

\begin{itemize}
  \item \textbf{UMass:} Uses asymmetric log conditional probability:
  \[
  M_{\text{UMass}}(w_i, w_j) = \log \frac{D(w_i, w_j) + \varepsilon}{D(w_j)},
  \]
  where \( D(w_i, w_j) \) counts co-occurrences in documents.
  
  \item \textbf{NPMI:} Normalized Pointwise Mutual Information:
  \[
  M_{\text{NPMI}}(w_p, w_q) = \frac{\log \frac{P(w_p, w_q)}{P(w_p) P(w_q)}}{-\log P(w_p, w_q)}.
  \]

  \item \textbf{\( C_{\!v} \):} Measures cosine similarity between NPMI vectors \( \mathbf{v}_i \) and \( \overline{\mathbf{v}} \):
  \[
  M_{C_{\!v}}(w_i, T \setminus \{w_i\}) = \cos(\mathbf{v}_i, \overline{\mathbf{v}}).
  \]
\end{itemize}

\paragraph{Aggregation \( \Sigma \).}
The Arithematic mean value of all pairwise confirmation ccores:
\[
Coherence(T) = \frac{1}{m} \sum_{i=1}^m M(W'_i, W^*_i; P).
\]

\medskip

This abstract formulation accommodates most coherence metrics by selecting appropriate design choices for \( S, P, M, \text{and }\Sigma \). For example:
\begin{itemize}
  \item \textbf{$C_{UMass}$}: asymmetric segmentation, document co-occurrence, log-probability, mean. The $C_{UMass}$ coherence is computed by:
  
\[
  C_{UMass}(T)=\frac{1}{\binom{N}{2}}\sum_{i<j}\log\frac{D(w_i,w_j)+\varepsilon}{D(w_j)},
\]
where $D(w_i,w_j)$ denotes the number of documents containing both $w_i$ and
$w_j$, $D(w_j)$ counts documents containing $w_j$, and $\varepsilon$ is a small
smoothing constant.\\
This score evaluates the topic with respect to a number of documents sharing the same words in the top words, or the term-document matrix in asymmetric conditional probabilities.
  
  \item \textbf{$C_{NPMI}$}: pairwise segmentation, sliding window estimation, PMI confirmation, mean. The $C_{NPMI}$ coherence is a Normalized Point‑wise Mutual Information for a pair of words $w_i,w_j$. It is defined as:
  \[
  M_{\text{NPMI}}(w_p, w_q) = \frac{\log \frac{P(w_p, w_q)}{P(w_p) P(w_q)}}{-\log P(w_p, w_q)}.
  \]
  This metric rewards word pairs whose joint probability exceeds what independence would predict, normalizing by self-information so that scores fall in $[-1,1]$ and remain comparable across corpora. Aggregating NPMI over the top word pairs yields a coherence score for the entire topic.
  
  \item \textbf{$C_{v}$}: one-set segmentation, external corpus with sliding window, cosine similarity, mean. The $C_{\!v}$ metric instantiates all four
dimensions of their \emph{coherence design space}---segmentation $(S)$, probability estimation $(P)$, confirmation measure $(M)$, and aggregation $(\Sigma)$---as follows:
\begin{itemize}
  \item \textbf{Segmentation} $S$: one–set segmentation $(S_{\text{one\_set}})$
  treats the first word $w_i$ as a singleton and the remainder of the top-$N$
  list as a complementary set, yielding ordered pairs $\langle\{w_i\},\{w_{j\neq i}\}\rangle$.
  \item \textbf{Probability} $P$: co‑occurrence counts are collected in a large
  sliding window of $110$ tokens over an external reference corpus (Wikipedia),
  denoted $P_{\mathrm{sw}}(110)$.  
  \item \textbf{Confirmation} $M$: for each pair the Normalized PMI values are
  assembled into confirmation vectors; semantic relatedness between two
  segments is then computed via cosine similarity of these vectors (an
  \emph{indirect} confirmation function).
  \item \textbf{Aggregation} $\Sigma$: the arithmetic mean of all pairwise
  confirmation scores provides the final topic‑level value.
\end{itemize}
This metric first embeds every word in a vector of its NPMI relations to the other words, then measures how similar these relation patterns are via cosine similarity. Formally, let $\mathbf{v}_i$ be the NPMI vector of word $w_i$ against the remaining set, then
 
\[
  C_{\!v}(T)=\frac{1}{N}\sum_{i=1}^{N}\cos\bigl(\mathbf{v}_i,\overline{\mathbf{v}}\,\bigr),
\]
where $\overline{\mathbf{v}}$ is the mean confirmation vector, and cosine is bounded in $[-1,1]$.  Thanks to the large window and indirect confirmation, $C_{\!v}$ achieves the highest average Pearson correlation ($r\!\approx\!0.73$) with human topic ratings across five gold‑standard datasets. Thus, it functions as the default coherence metric in common-used python libraries like \texttt{BERTopic} \cite{grootendorst2022bertopic} and \texttt{Gensim} \cite{rehurek_lrec}. 
Moreover, the $C_{v}$ metric, was identified as the best performer in the exhaustive study by \cite{roder2015exploring}. 

\end{itemize}

In the discussions which follow, this metric shall be used to compute topic coherence.

\subsubsection{Topic Divergence}\label{subsec:topic_divergence_formal}
While topic coherence focuses on whether the top words that define each topic actually go together, \textbf{topic divergence} is the statistic that quantifies the dissimilarity between several themes. It is undesirable for a topic model to produce duplicate and overlapping topics. If two topics have the same set of highly probable words, they may be representing the same meaning, which makes interpretation of topics difficult.

High topic divergence shows that learned topics are capturing different parts of the corpus. This quality is important for applications such as document classification, thematic clustering or exploratory data analysis. Measuring this helps evaluate if there are enough topics for the diveristy of the corpus that has been used for training, whether the model is generating semantically diverse topics or output generated by the model represents too much similar content.

\textbf{Definition 2.2 Topic Divergence}:\label{def:topic_divergence}
Let \( \Phi = \{\phi_1, \phi_2, \dots, \phi_K\} \) denote the set of \( K \) topic-word distributions learned by the model, where each \( \phi_k \in \Delta^V \) is a probability vector over a vocabulary of size \( V \), i.e.,
\[
\phi_k = [\phi_k(w_1), \dots, \phi_k(w_V)], \quad \text{with} \quad \sum_{j=1}^V \phi_k(w_j) = 1.
\]

Let \( D(\cdot \parallel \cdot) \) be a divergence or distance on probability vectors. The \textbf{average pairwise topic divergence} is defined by:

\[
\text{TopicDivergence}(\Phi) = \frac{2}{K(K-1)} \sum_{1 \leq i < j \leq K} D(\phi_i \parallel \phi_j)
\]

Larger values indicate that topics are more clearly separated from one another.

\paragraph{Common Divergence Measures.}
\begin{itemize}

\item \textbf{Jensen--Shannon Divergence (JSD)} 
is a symmetrized and smoothed version of the Kullback--Leibler (KL) divergence. For discrete distributions  \( \phi_i \) and \( \phi_j \) on the same vocabulary and their mixture $M = \tfrac{1}{2}(\phi_i + \phi_j)$, one has: 
    \[
    \text{JSD}(\phi_i \parallel \phi_j) = \frac{1}{2} D_{\mathrm{KL}}(\phi_i \parallel M) + \frac{1}{2} D_{\mathrm{KL}}(\phi_j \parallel M),
    \]
    where \( D_{\mathrm{KL}} \) denotes KL divergence. With base‑2 logarithms, $\text{JSD}\in[0,1]$.  
JSD is commonly used to maximize inter‑topic separation when choosing the number of topics~\cite{deveaud2014}.

    \item \textbf{Hellinger Distance:} A geometric measure that computes the \( \ell^2 \) distance between the square roots of two probability vectors:
    \[
    D_H(\phi_i, \phi_j) = \frac{1}{\sqrt{2}} \left\| \sqrt{\phi_i} - \sqrt{\phi_j} \right\|_2.
    \]
Empirical studies using hierarchical clustering under the Hellinger distance (e.g.~\cite{uon2021hellinger}) reveal high‑level thematic structure in very large topic models.

    \item \textbf{Cosine Distance:} an angle‑based dissimilarity (one minus cosine similarity):
    \[
    D_{\text{cos}}(\phi_i, \phi_j) = 1 - \frac{\phi_i \cdot \phi_j}{\|\phi_i\| \|\phi_j\|}.
    \]    
\end{itemize}

Topic divergence is widely used to identify redundant or overlapping topics, to select $K$ by maximizing inter‑topic separation, and to visualize the topic geometry via clustering or dimensionality‑reduction methods.




\section{Materials and Methods}~\label{sec:methods}

In this study we examine how the size of the underlying transformer‑based language model affects topic decomposition in the BERTopic \cite{grootendorst2022bertopic} pipeline. Our aim is to test whether scaling improves topic coherence and increases topic diversity. To that end, we run a controlled set of comparisons across several transformer encoders within one unified pipeline and evaluate them using standardized metrics. This section details the experimental setup, covering datasets, preprocessing, modeling pipeline, and the encoders under comparison.

\subsection{Description of the Datasets}

We conduct our experiments on eleven distinct text corpora:

\paragraph{20 Newsgroups.} 
The \textit{20 Newsgroups} corpus \cite{twenty_newsgroups_113} is a widely used benchmark of about twenty thousand documents organized into 20 categories. Its topics span politics, sports, religion, science, and more, providing a heterogeneous testbed for topic modeling and a standard baseline for assessing coherence and alignment with human labels.

\paragraph{AG News Corpus.} The AG News Corpus contains approximately 120,000 news articles collected from online sources in four categories: World, Sports, Business, and Sci/Tech. A widely used benchmark for text classification models, including deep learning baselines. The data are available in CSV format with columns for class index, title, and description.

\paragraph{Amazon Reviews (2018).}For a structured consumer-review domain, we employ the 2018 Amazon Reviews dataset, which contains large-scale product reviews with ratings, detailed metadata, and category labels. Each review includes fields such as \texttt{reviewText}, \texttt{summary}, \texttt{overall} rating, and product identifiers, forming a rich benchmark for sentiment analysis, aspect extraction, and recommendation-related topic modeling. Its breadth across product categories and mixture of narrative and structured data provides a controlled environment for studying semantic alignment and domain adaptation.

\paragraph{BBC News Dataset.} The BBC News Dataset contains 2,225 news articles published by BBC grouped into five categories: business, entertainment, politics, sport, and tech. Used for basic text classification and clustering experiments due to its clean labeling and balanced topics. The data are available as plain text files organized in folders by category.

\paragraph{CORD-19 Dataset}This dataset contains many academic papers about COVID-19 Research, with the latest release being 18.7 GB in size. The data is mostly contained in .tar files, but for our purposes, we use the file \texttt{metadata.csv} given with the dataset to access the abstracts for the papers. The dataset can be downloaded at \url{https://github.com/allenai/cord19}. For this study, we sampled 10000 abstracts from the dataset.

\paragraph{IMDb Large Movie Review Dataset.}We also incorporate the IMDb movie review corpus, a long-standing benchmark for text sentiment classification. It consists of 25{,}000 positive and 25{,}000 negative labeled reviews, with an additional unlabeled subset for semi-supervised experiments. As a clean and balanced dataset focused on opinion-rich text, IMDb offers a stable baseline for evaluating coherence metrics, sentiment-topic disentanglement, and embedding quality on short to medium-length narratives.

\paragraph{PubMed Collection.}
We curated a domain‑specific set of recent PubMed abstracts to complement our general-domain corpora. Using a custom Python script\footnote{Included in our code repository.} and the \textit{NCBI Entrez API}, we retrieved papers dated between Jan 1, 2024 and May 31, 2025 containing the terms \texttt{"Large Language Models"}, \texttt{"AI‑assisted diagnosis"}, or \texttt{"Artificial Intelligence"}. The resulting CSV dataset\footnote{Available at: \url{https://huggingface.co/datasets/willyrv/pubmed2024_IA-LLM-diagnostic}} includes roughly 49.2k English abstracts with metadata (PMID, title, authors, journal, date, DOI, keywords, links) and is released under MIT. It captures current trends in biomedical AI and is well suited for evaluating topic models in a domain‑centric setting.

\paragraph{Pushshift Reddit Collection.}To incorporate a large-scale, real-world social discourse dataset, we utilize a curated slice of the \textit{Pushshift Reddit} archives. The collection contains historical Reddit posts and comments stored in JSON format, spanning multiple years and covering a broad range of communities. Its scale---hundreds of millions of entries in full form---captures evolving online conversations, polarization patterns, and community-specific jargon. This dataset enables evaluation of topic coherence, temporal topic drift, and robustness of clustering methods under noisy, user-generated content.

\paragraph{Reuters-21578} The Reuters-21578 corpus contains 21,578 Reuters financial and news articles from 1987 manually labeled by topic categories such as “earn,” “acq,” and “trade.” Used for text categorization, topic modeling, and document classification benchmarking. The data are available as SGML or plain text files with topic labels in metadata.

\paragraph{Wikipedia Abstracts} This dataset contains abstracts from many Wikipedia articles, with the English dataset having 6.58 million rows. More details may be found at \url{https://huggingface.co/datasets/laion/Wikipedia-Abstract}. In this study, we sampled 10000 abstracts from the English subset.

\paragraph{Yahoo Answers Topics.} The Yahoo Answers Topics corpus is a large corpus of community-generated questions and answers from Yahoo Answers with 10 main categories. Each record includes a question title, content, best answer, and topic label. Commonly used for topic classification, question-answer modeling, and text summarization research. The data are available in CSV or JSON format.

The versions of the above described datasets that were used on our experiments can be downloaded here: \url{https://huggingface.co/datasets/willyrv/text_datasets_for_experiments/tree/main}.

\subsection{Preprocessing steps}\label{subsec:preprocessing}

Before running BERTopic, we apply a small set of preprocessing steps to standardize and normalize inputs. The goal is to reduce noise, improve embedding quality, and keep processing consistent across datasets.

\paragraph{Tokenization.}
The first step in preprocessing a document is to \textbf{tokenize} it—i.e., break it down into individual words or tokens. In our pipeline, we use the \texttt{simple\_preprocess} function from the \texttt{gensim} library. It performs whitespace tokenization and additionally removes punctuation, sets all characters to lowercase, and filters out short words. While this is a lightweight approach, it ensures speed and compatibility with the downstream topic modeling pipeline.

\paragraph{Stopword Removal.}
During tokenization, common \textbf{stop words}—i.e., function words that carry little semantic content such as ``the", ``and", ``but", etc.—are automatically filtered out by \texttt{simple\_preprocess}. This improves the focus of the model on content-rich terms that are more indicative of latent topics.

\paragraph{Lemmatization.}
While full lemmatization (i.e., reducing words to their root forms such as ``better'' $\rightarrow$ ``good'' or ``went'' $\rightarrow$ ``go'') is not explicitly performed in the current implementation, the lowercase filtering and punctuation removal from \texttt{simple\_preprocess} contribute to partial normalization of word forms. In future extensions, integration with a lemmatization toolkit such as spaCy~\cite{honnibal2020spacy} could improve linguistic normalization further.

\paragraph{Corpus Preparation.}
After preprocessing, we build a bag‑of‑words representation using \texttt{gensim}'s \texttt{Dictionary}, which assigns a unique ID to each token. Documents are then converted to sparse vectors via \texttt{doc2bow} to record per‑document token frequencies. These structures enable efficient computation of coherence metrics.

\paragraph{Motivation.}
This preprocessing pipeline balances simplicity and effectiveness: it standardizes input across domains (general and biomedical), filters irrelevant noise, and generates the linguistic units necessary for calculating both coherence and diversity scores. It ensures that the input to the transformer-based embedding models reflects meaningful semantic structure.

\subsection{The BERTopic pipeline}\label{subsec:bertopic}

To perform topic decomposition, we rely on the BERTopic framework~\cite{grootendorst2022bertopic}, a modern and flexible topic modeling algorithm that leverages transformer-based embeddings to extract semantically meaningful topics. BERTopic proceeds in three main stages:

\begin{enumerate}
    \item \textbf{Document Embeddings.} Each preprocessed document is embedded into a high-dimensional vector space using a pre-trained transformer-based language model. These embeddings preserve semantic relationships, which account for different contexts, between words and documents. BERTopic is model-agnostic at this stage, allowing any sentence embedding model compatible with the \texttt{sentence-transformers} library~\cite{reimers-2019-sentence-bert}.
    
    \item \textbf{Clustering.} The document embeddings are then clustered using a density-based algorithm—typically HDBSCAN~\cite{campello2013density}—which groups semantically similar documents without requiring a fixed number of clusters. Each resulting cluster is treated as a potential topic.
    
    \item \textbf{Topic Representation.} For each cluster, BERTopic uses a class-based variant of TF-IDF to identify the most representative words. This approach (referred to as \textit{c-TF-IDF}) computes the importance of a word in a cluster relative to the entire corpus, providing interpretable keyword lists for each discovered topic.
\end{enumerate}

\paragraph{Model Variation.}
In our experiments, we leverage BERTopic’s modular design to substitute the underlying embedding model and evaluate how this choice affects the coherence and diversity of the extracted topics. Specifically, we test a range of transformer-based models of varying sizes (e.g., MiniLM, DistilBERT, BERT-base, and larger LLaMA variants). This allows us to isolate the influence of model size (measured in number of parameters) on the interpretability and discriminability of topics. The number of topic that was used for each model was decided by the "default" internal parameters of BERTopic\cite{bertopic_best_practices}, via clustering algorithm we mentioned earlier, and the metrics could be slightly different if we change that meta-parameter.

\paragraph{Advantages.}
The ability to switch embedding models without modifying the rest of the pipeline makes BERTopic particularly well-suited for comparative evaluations. Moreover, the use of context-aware embeddings improves the handling of polysemous words and enhances the semantic quality of the topics.

\subsection{Embedding Models Used For Comparison}\label{subsec:models}

We evaluated seven transformer-based embedding models spanning a range of parameter scales, from compact (22 Millions) to extremely large (13 Billions), to explore how model size affects topic quality within the BERTopic pipeline.

\paragraph{\textbf{all‑MiniLM‑L6‑v2}.}  
A compact Sentence‑Transformer with only \textbf{$\sim$22 M parameters} designed specifically for sentence embedding tasks. It transforms sentences into 384-dimensional vectors using a contrastive training objective over more than one billion pairs. Despite its small size, it delivers high-quality semantic embeddings ideal for clustering and topic modeling. More details may be found at https://huggingface.co/sentence-transformers/all-MiniLM-L6-v2.

\paragraph{\textbf{MiniLM‑L12‑H384‑uncased} \cite{wang2020minilm}.}  
MiniLM retains the structural complexity of deeper models (12 transformer layers and 384 hidden dimensions) but reduces parameter count to around \textbf{33 M}, striking a balance between performance and efficiency.

\paragraph{\textbf{DistilBERT} \cite{Sanh2019DistilBERT}.}  
A distilled version of BERT‑Base with \textbf{$\sim$66 M parameters} ($\sim$40\% fewer than BERT\textsubscript{BASE}) achieved by reducing depth from 12 to 6 layers and employing knowledge distillation alongside a masked‑language modeling objective. DistilBERT preserves $\sim$97\% of BERT’s performance while being significantly faster and smaller.

\paragraph{\textbf{BERT‑base‑uncased} \cite{devlin2019bert}.}  
The original BERT‑Base model with \textbf{110 M parameters}, featuring 12 encoder layers and 768 hidden units. It was pretrained on over 3.3 B tokens combining \textit{BooksCorpus} and English Wikipedia materials, using masked language modeling and next-sentence prediction.

\paragraph{\textbf{RoBERTa‑base} \cite{liu2019roberta}.}  
An optimized variant of BERT with approximately \textbf{125 M parameters} that removes the next-sentence prediction and introduces dynamic masking. Trained on a much larger corpus ($\sim$160 GB text including \textit{CC-News} and \textit{OpenWebText}), RoBERTa improves learning efficiency and downstream performance.

\paragraph{\textbf{LLaMA‑2‑7B} \cite{touvron2023llama}.}  
A decoder-only LLM with \textbf{$\sim$7 B parameters}, trained on 2 trillion tokens and employing grouped-query attention to optimize inference. The model offers strong generative capabilities and represents the low end of the large-model spectrum.

\paragraph{\textbf{LLaMA‑2‑13B}.}  
The medium-sized version of the family with \textbf{$\sim$13 B parameters}, pretrained similarly on massive datasets and achieving benchmark performance competitive with 175 B‑parameter models.

\subsection{Model Size and Memory Constraints}
Transformer-based language models with billions of parameters—such as \texttt{LLaMA-2-13B}\\—require substantial GPU memory, often exceeding 24--32 GB even for inference. Each parameter in a model is typically stored as a 32-bit floating point value (FP32), and during inference, intermediate activations and gradient caches also consume significant memory. As a result, deploying such large-scale models on commodity hardware (e.g., single GPUs or limited CPU RAM) becomes infeasible without optimization.

\paragraph{Quantization for Efficient Inference}
To address these limitations, we apply \textbf{quantization}, a compression technique that reduces the bit-width of the model parameters. In our experiments, we use \emph{8-bit quantization}, storing each weight with 8 bits instead of 32. This reduces memory usage by up to 75\% and enables inference of large models like \texttt{LLaMA-2-13B} on modest hardware configurations. Quantization trades off some numerical precision for large gains in efficiency, often with negligible impact on downstream performance—especially for embedding or topic modeling tasks that are less sensitive to slight variations in representation. It is a widely adopted technique in model deployment pipelines for reducing latency and resource demands, particularly in production environments or academic setups with constrained compute resources.

\paragraph{Reproducibility}
The full experimental code, including data collection, preprocessing, and evaluation scripts, can be found at \cite{ml2nlp-code}.

\section{Results}~\label{sec:results}
In the following section, we present the results of our comparative evaluation of topic models generated by BERTopic using various transformer-based language models. The experiments were carried out on a workstation using Ubuntu 24.04.3 LTS with kernel 6.14.0-28-generic. The system features an AMD Ryzen 7 7700X CPU (8 cores, 16 threads), 64 GB of RAM, and an NVIDIA GeForce RTX 4090 GPU with 24 GB of memory (CUDA 12.9, driver 575.64.03).

\begin{figure}
    \centering
    \includegraphics[width=\linewidth]{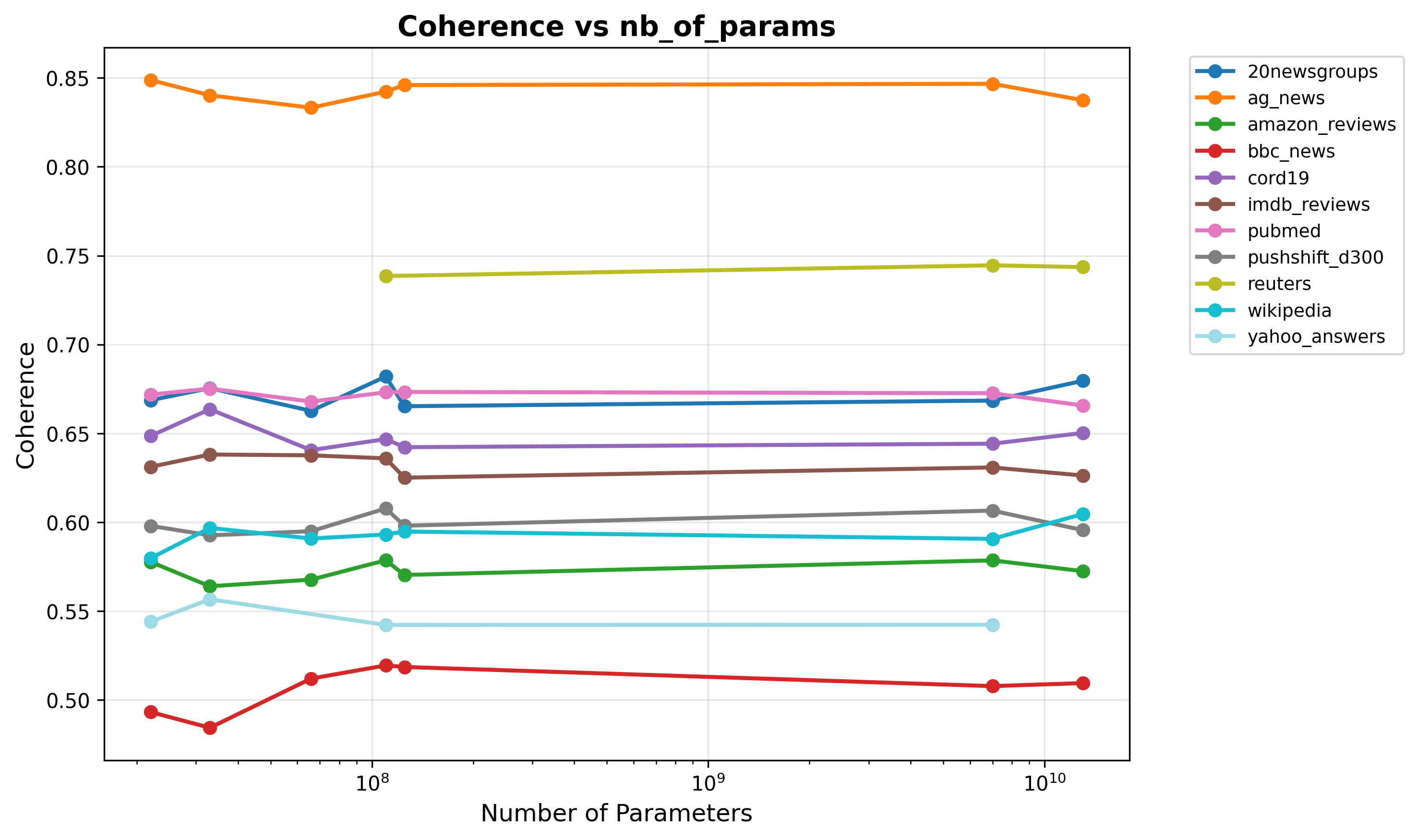}
    \caption{Topic coherence across datasets as a function of model size (parameters). Each line represents a different dataset.}
    \label{fig:coherencevsnb_of_params}
\end{figure}

\begin{figure}
    \centering
    \includegraphics[width=1\linewidth]{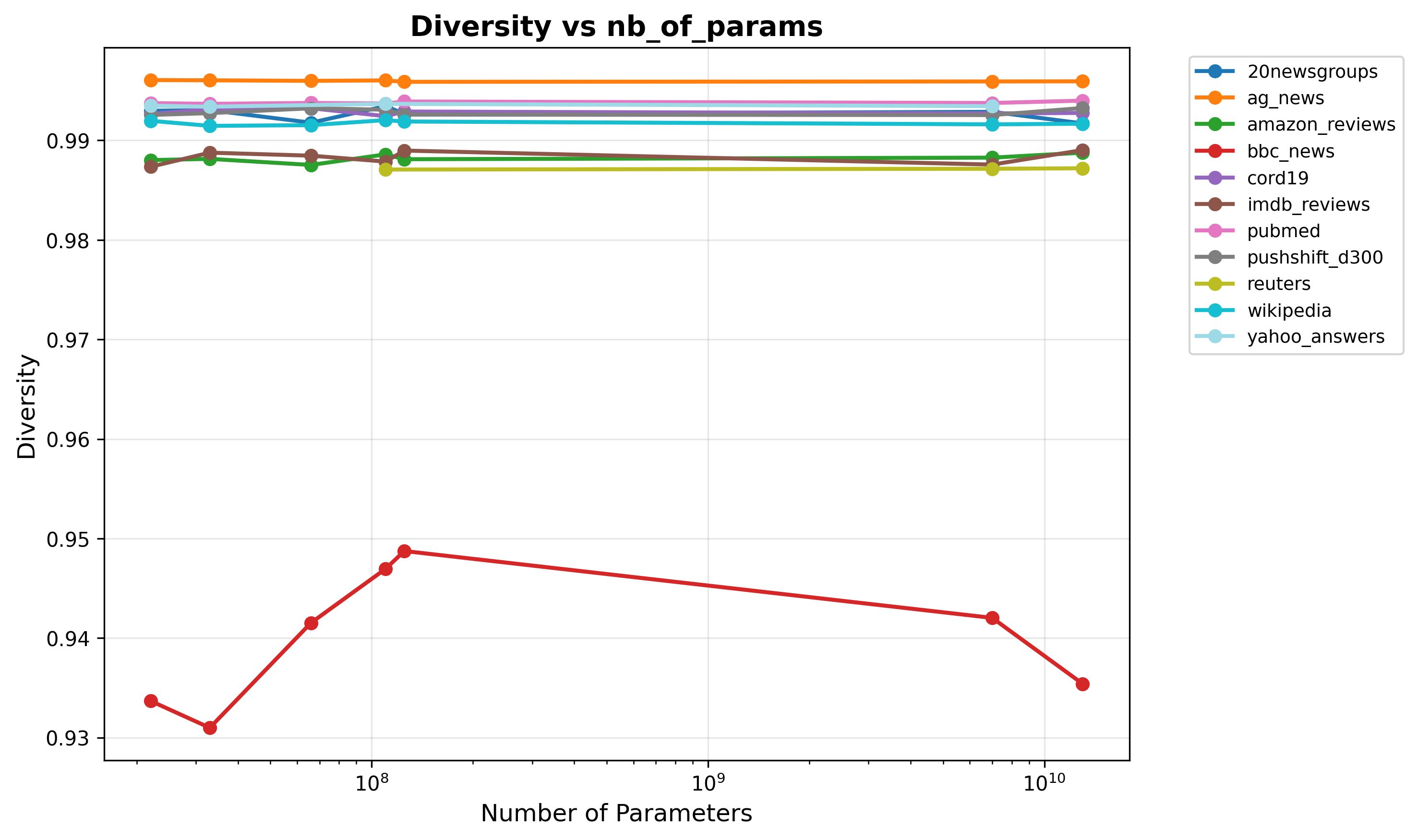}
    \caption{Topic diversity across datasets as a function of model size (parameters). Each line represents a different dataset.}
    \label{fig:diversityvsnb_of_params}
\end{figure}

\begin{figure}[H]
\centering
\begin{tikzpicture}
\begin{axis}[
    width=13cm,
    height=8cm,
    xmode=log,
    xlabel={Model Size (parameters, log scale)},
    ylabel={Mean Topic Coherence (across datasets)},
    title={Model Size vs Mean Topic Coherence (11 datasets)},
    legend columns=1,
    legend style={at={(0.98,0.02)}, anchor=south east, fill=white, draw=black},
    xmajorgrids=false,
    ymajorgrids=true,
    ymin=0, ymax=1,
    enlargelimits=0.08,
    error bars/.cd,
]

\addplot+[
    only marks,
    mark=o,
    color=black,
    error bars/.cd,
      y dir=both,
      y explicit,
] coordinates {
    (22000000,0.6165) +- (0,0.1042)
    (33000000,0.6194) +- (0,0.1040)
    (66000000,0.6234) +- (0,0.1003)
    (110000000,0.6321) +- (0,0.1011)
    (125000000,0.6419) +- (0,0.0971)
};
\addlegendentry{Non-Quantized (small models)}

\addplot+[
    only marks,
    mark=triangle*,
    color=black,
    error bars/.cd,
      y dir=both,
      y explicit,
] coordinates {
    (7000000000,0.6427) +- (0,0.0993)
    (13000000000,0.6544) +- (0,0.0957)
};
\addlegendentry{Quantized (Llama)}

\end{axis}
\end{tikzpicture}
\begin{center}
\caption{Mean topic coherence versus model size (log scale). Each point represents an endocer, and error bars show one standard deviation across datasets.
}  
\label{fig:model_size_vs_mean_coherence}
\end{center}
\end{figure}
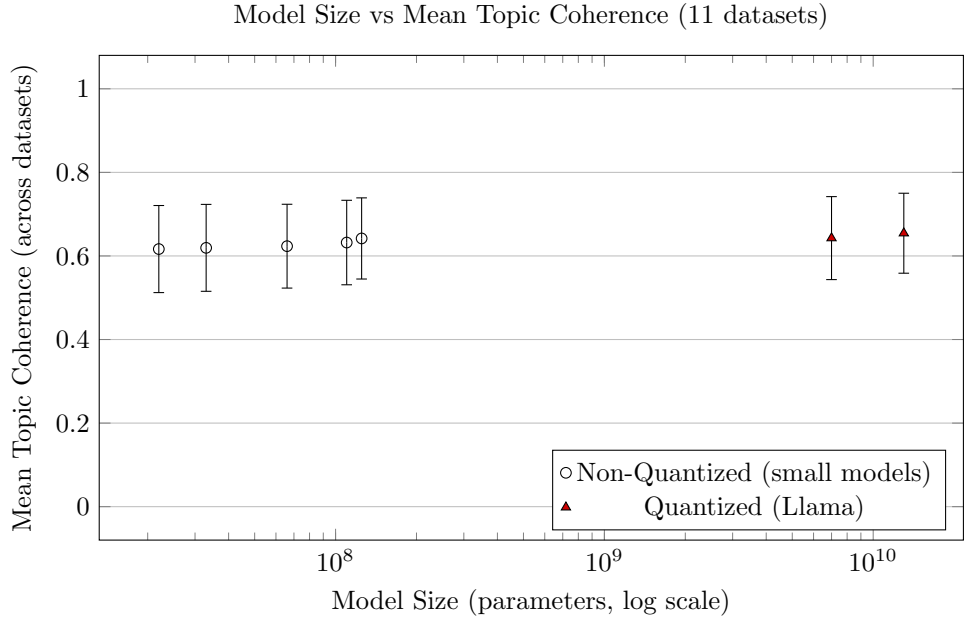

Our primary goal was to assess how the size of the embedding model influence the coherence and diversity of discovered topics. We benchmarked seven encoders, ranging from compact models like MiniLM and DistilBERT to large-scale architectures such as RoBERTa and LLaMA-2 (7B and 13B). Each model was tested both in its original form and under 8-bit quantization, allowing us to explore performance trade-offs under memory-constrained settings. We report two primary metrics: topic coherence (which measures the semantic consistency of top words within each topic) and topic diversity (which reflects inter-topic dissimilarity and redundancy).
Note that for the LLaMA-2 13B model, we only report results under quantization. Due to its large memory footprint, the non-quantized version of LLaMA-2 13B could not be executed within our hardware constraints. Consequently, its row appears empty in the unquantized results table, and only its quantized scores are available for comparison.
The performance of the models on each corpus are illustrated in Figure~\ref{fig:coherencevsnb_of_params} and Figure~\ref{fig:diversityvsnb_of_params}. The mean coherence values achieved by each model across all corpora are illustrated in Figure~\ref{fig:model_size_vs_mean_coherence}.


\begin{table}[H]
\centering
\small
\setlength{\tabcolsep}{3.5pt}
\renewcommand{\arraystretch}{1.15}
\resizebox{\textwidth}{!}{%
\begin{tabular}{|l|c|c|c|c|c|c|c|}
\hline
\textbf{Dataset} & \textbf{MiniLM-L6} & \textbf{MiniLM-L12} & \textbf{DistilBERT} & \textbf{BERT-base} & \textbf{RoBERTa} & \textbf{Llama2-7B (Q)} & \textbf{Llama2-13B (Q)} \\
\hline
20Newsgroups & 0.6687 & 0.6754 & 0.6627 & \textbf{0.6820} & 0.6653 & 0.6685 & 0.6796 \\
\hline
AG News & \textbf{0.8488} & 0.8402 & 0.8332 & 0.8423 & 0.8460 & 0.8467 & 0.8374 \\
\hline
Amazon Reviews & 0.5777 & 0.5640 & 0.5677 & \textbf{0.5786} & 0.5704 & 0.5786 & 0.5725 \\
\hline
BBC News & 0.4932 & 0.4844 & 0.5121 & \textbf{0.5194} & 0.5186 & 0.5078 & 0.5095 \\
\hline
CORD-19 & 0.6486 & \textbf{0.6636} & 0.6406 & 0.6468 & 0.6422 & 0.6442 & 0.6503 \\
\hline
IMDb & 0.6312 & \textbf{0.6381} & 0.6377 & 0.6360 & 0.6251 & 0.6308 & 0.6263 \\
\hline
PubMed & 0.6718 & \textbf{0.6752} & 0.6679 & 0.6731 & 0.6733 & 0.6741 & 0.6654 \\
\hline
Pushshift & 0.5980 & 0.5926 & 0.5950 & \textbf{0.6078} & 0.5981 & 0.6066 & 0.5956 \\
\hline
Reuters & \textemdash & \textemdash & \textemdash & 0.7385 & \textemdash & \textbf{0.7445} & 0.7436 \\
\hline
Wikipedia & 0.5798 & \textbf{0.5968} & 0.5909 & 0.5931 & 0.5948 & 0.5928 & 0.5888 \\
\hline
Yahoo Answers & 0.5440 & \textbf{0.5567} & \textemdash & 0.5422 & \textemdash & 0.5423 & \textemdash \\
\hline
\end{tabular}}
\caption{\textbf{Topic coherence across 11 datasets.}
Bold indicates the best encoder per dataset; (Q) denotes quantized runs for Llama-2. Missing entries were not evaluated.}
\label{tab:coherence_matrix_11}
\end{table}

\begin{table}[H]
\centering
\small
\setlength{\tabcolsep}{3.5pt}
\renewcommand{\arraystretch}{1.15}
\resizebox{\textwidth}{!}{%
\begin{tabular}{|l|c|c|c|c|c|c|c|}
\hline
\textbf{Dataset} & \textbf{MiniLM-L6} & \textbf{MiniLM-L12} & \textbf{DistilBERT} & \textbf{BERT-base} & \textbf{RoBERTa} & \textbf{Llama2-7B (Q)} & \textbf{Llama2-13B (Q)} \\
\hline
20Newsgroups & 0.9930 & 0.9930 & 0.9918 & \textbf{0.9934} & 0.9927 & 0.9929 & 0.9917 \\
\hline
AG News & \textbf{0.9961} & 0.9960 & 0.9960 & 0.9960 & 0.9959 & 0.9959 & 0.9959 \\
\hline
Amazon Reviews & 0.9880 & 0.9881 & 0.9875 & 0.9886 & 0.9881 & 0.9883 & \textbf{0.9888} \\
\hline
BBC News & 0.9337 & 0.9310 & 0.9415 & 0.9470 & \textbf{0.9487} & 0.9420 & 0.9354 \\
\hline
CORD-19 & 0.9926 & 0.9932 & \textbf{0.9932} & 0.9924 & 0.9929 & 0.9927 & 0.9928 \\
\hline
IMDb & 0.9946 & 0.9947 & \textbf{0.9954} & 0.9946 & 0.9945 & 0.9946 & 0.9941 \\
\hline
PubMed & 0.9941 & 0.9941 & 0.9942 & \textbf{0.9943} & 0.9940 & 0.9942 & 0.9941 \\
\hline
Pushshift & 0.9926 & 0.9927 & 0.9932 & 0.9931 & 0.9926 & 0.9925 & \textbf{0.9933} \\
\hline
Reuters & \textemdash & \textemdash & \textemdash & 0.9871 & \textemdash & 0.9871 & \textbf{0.9872} \\
\hline
Wikipedia & \textbf{0.9936} & 0.9934 & 0.9932 & 0.9931 & 0.9933 & 0.9932 & 0.9929 \\
\hline
Yahoo Answers & 0.9935 & 0.9934 & \textemdash & \textbf{0.9937} & \textemdash & 0.9934 & \textemdash \\
\hline
\end{tabular}}
\caption{\textbf{Topic diversity across 11 datasets.} Bold indicates the best encoder per dataset; (Q) denotes quantized runs for Llama-2. Missing entries were not evaluated.}
\label{tab:diversity_matrix_11}
\end{table}

\section{Conclusions}~\label{sec:conclusions}

Contrary to intuitive expectations, our empirical results suggest that increasing the complexity of transformer-based language models has little to no impact on the quality of the topics discovered in unsupervised topic modeling.
Experiments were carried out with the BERTopic  \cite{grootendorst2022bertopic} framework, which employs a contemporary pipeline with document embeddings and clustering to produce meaningful and distinct topic arrangements. We propose an approach to measure the quality of the topics through systematic variation of the embedding model, ranging from small architectures such as MiniLM to large-scale models such as LLaMA-2, which consists of up to 13 billion parameters.

To quantify the interpretability of the generated topics, we relied on a set of established automatic metrics for topic coherence, grounded in the theoretical framework proposed by \cite{roder2015exploring}. This framework decomposes coherence evaluation into four components: segmentation, probability estimation, confirmation measure, and aggregation function. By adopting metrics such as $C_v$, we ensured a robust and multifaceted assessment of both intra-topic consistency and inter-topic divergence.

The pipeline we implemented builds on BERTopic and introduces a modular design that facilitates the replacement and comparison of different embedding models while keeping all downstream processing stages (e.g., dimensionality reduction, clustering, topic extraction) fixed. This allowed us to isolate the influence of the embedding space on the quality of the topics, as measured by coherence and diversity metrics.

Our results, drawn from both general-domain (20 Newsgroups) and specialized-domain (PubMed abstracts) datasets, indicate that increasing the size of the transformer encoder does not necessarily improve topic interpretability. In fact, smaller models such as DistilBERT and MiniLM performed comparably to their larger counterparts. Similarly, applying 8-bit quantization had negligible effects on the coherence and diversity of the resulting topics, suggesting that model efficiency can be significantly improved without sacrificing interpretability.

Although the present findings are promising, the study itself is not definitive nor all-encompassing. The findings need to be generalized by testing in other kinds of text corpora, including not only non-English data, but also multi-domain corpora and low-resource corpora. Moreover, further studies could try to explore more kinds of models, embedding techniques, and training objectives, especially the ones which were adapted for specific sentence-level semantical task or dense retrieval-related tasks. In this way, we could have further understanding on the relationships between model complexity, computational cost and interpretability.

In conclusion, our research demonstrates the strong methodological approach of controlled experiments while firstly revealing an unexpected yet useful finding that larger models do not always produce better results. The DistilBERT model with 66M parameters achieves topic coherence similar to that of LLaMA-7B with 7B parameters while requiring less than 1 percent of memory and running faster. The efficiency-effectiveness trade-off makes lightweight models highly attractive for practitioners seeking scalable and interpretable topic modeling solutions, which is crucial for resource-limited deployments.

\vskip 0.4 cm

\noindent{\bf Acknowledgments}\vskip 0.4 cm

The authors sincerely thank the MIT PRIMES program for providing a unique research experience that enabled this collaboration and the completion of this work. Also, the authors would like to thank the company Torus AI for generously providing the computational resources necessary to carry out the experiments.


\newpage
\printbibliography
\end{document}